\documentclass[pdflatex,iicol,sn-mathphys-num]{sn-jnl}% Math and Physical Sciences Numbered Reference Style
\usepackage{graphicx}%
\usepackage{multirow}%
\usepackage{amsmath,amssymb,amsfonts}%
\usepackage{amsthm}%
\usepackage{mathrsfs}%
\usepackage[title]{appendix}%
\usepackage{xcolor}%
\usepackage{textcomp}%
\usepackage{manyfoot}%
\usepackage{booktabs}%
\usepackage{algorithm}%
\usepackage{algorithmicx}%
\usepackage{algpseudocode}%
\usepackage{listings}%
\theoremstyle{thmstyleone}%
\theoremstyle{thmstyletwo}%

\theoremstyle{thmstylethree}%

\begin{document}

\title[Article Title]{Advancing Aesthetic Image Generation via Composition Transfer}


\author[1]{\fnm{Kai} \sur{Zou}}\email{kzou@mail.ustc.edu.cn}
\author[2]{\fnm{Zhiwei} \sur{Zhao}}\email{zhiweizhao@hfut.edu.cn}
\author*[1]{\fnm{Bin} \sur{Liu}}\email{flowice@ustc.edu.cn}
\author[1]{\fnm{Nenghai} \sur{Yu}}\email{ynh@ustc.edu.cn}

\affil*[1]{\orgdiv{School of Cyber Science and Technology},
           \orgname{University of Science and Technology of China},
           \orgdiv{Anhui Province Key Laboratory of Digital Security},
           \orgaddress{\street{96 Jinzhai Road}, \city{Hefei},
                       \postcode{230026}, \state{Anhui}, \country{China}}}

\affil[2]{\orgdiv{School of Computer Science and Information Engineering},
          \orgname{Hefei University of Technology},
          \orgaddress{\street{193 Tunxi Road}, \city{Hefei},
                      \postcode{230009}, \state{Anhui}, \country{China}}}


\abstract{Composition is a cornerstone of visual aesthetics, influencing the appeal of an image. While its principles operate independently of specific content, in practice, composition is often coupled with semantics. As a result, existing methods often enhance composition either through implicit learning or by semantics-based layout control, rather than explicitly modeling composition itself. To address this gap, we introduce \textbf{Composer}, a framework rooted in aesthetic theory, designed to model composition in a semantic-agnostic manner. First, it supports \textbf{composition transfer} by extracting key composition-aware representations from a reference image and leveraging a tailored conditional guidance module to control composition based on pre-trained diffusion models. Second, when users specify only text themes without a composition reference, Composer supports theme-driven composition retrieval by leveraging the in-context learning capabilities of Large Vision-Language Models (LVLMs), achieving explicit \textbf{composition planning}. To enhance composition in a reference-free mode, we conduct text-to-composition fine-tuning on the trained control module to enable implicit composition planning. Furthermore, we curated a high-quality dataset comprising 2 million image-text pairs using state-of-the-art generative models to support model training. Experimental results demonstrate that Composer significantly enhances aesthetic quality in text-to-image tasks and facilitates personalized composition control and transfer, offering users precision and flexibility in the creative process.}




\maketitle
\section{Introduction}
\label{sec:intro}
\begin{figure}[h!t]
  \centering
   \includegraphics[width=1.0\linewidth]{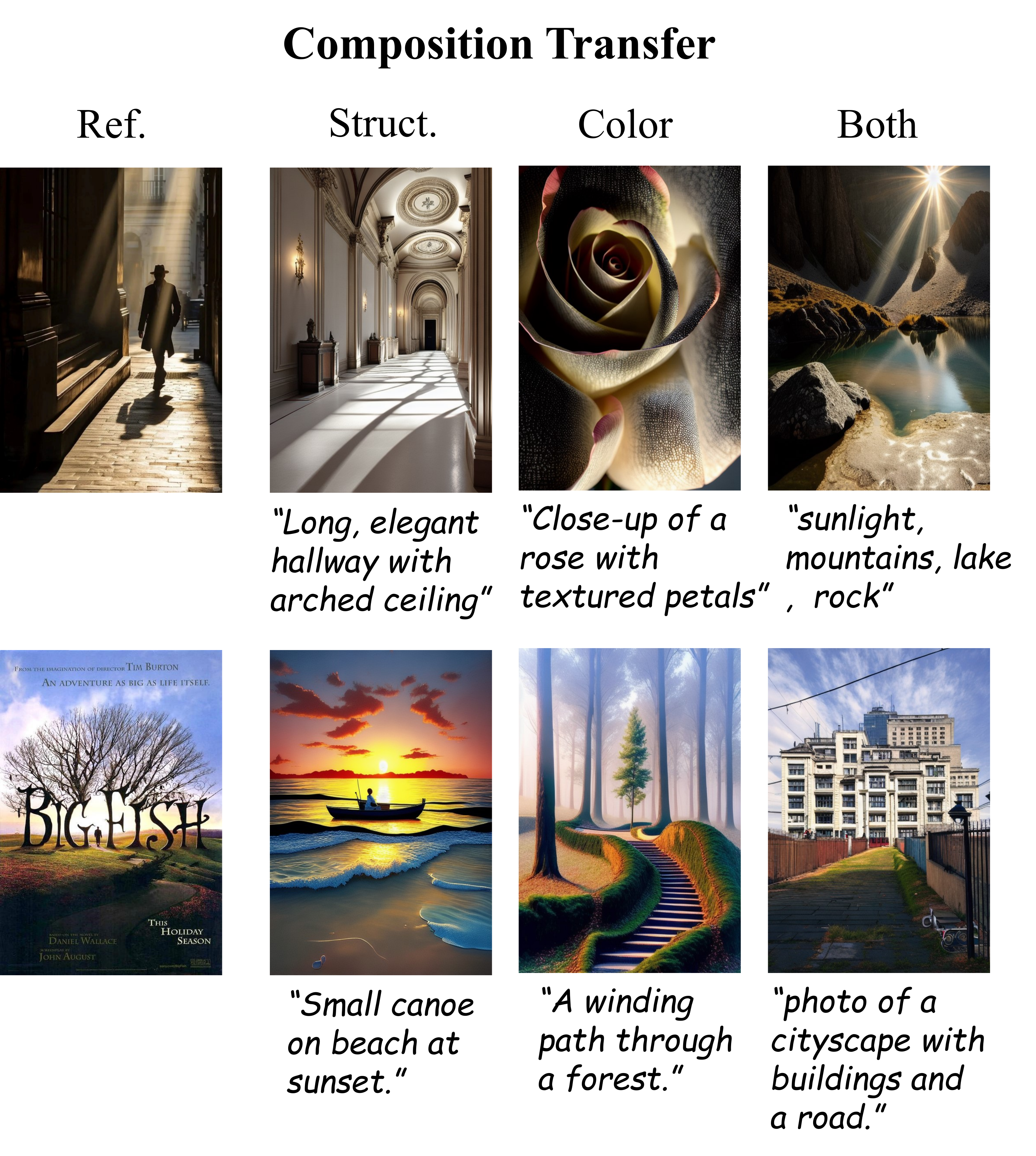}

   \caption{We present \textbf{Composer}, the first framework that supports explicit composition control, enabling composition transfer in image generation.
   From left to right: the reference image (\textbf{Ref.}), the result guided by spatial structure only (\textbf{Struct. only}), 
   the result guided by color distribution only (\textbf{Color only}), and the overall composition transfer result (\textbf{Both}).}
   \label{fig:1img}
\end{figure}
% \begin{figure}[ht]
%   \centering
%    \includegraphics[width=1\linewidth]{teaser.pdf}

%    \caption{
%     \textbf{Upper:} Compared to baseline(SD-v15), the proposed framework, \textbf{Composer}, demonstrates superior aesthetic composition while maintaining alignment with the given textual semantics. 
%     \textbf{Lower:} The framework also supports customized condition, allowing for spatial structure control through doodling or regional selection of color distribution using masks, providing flexible and precise control.
%     }
%    \label{fig:teaser}
% \end{figure}
Composition is the silent language of visual art, guiding the viewer's eye and shaping the emotional impact of an image~\cite{martin1983power}. A well-composed image weaves spatial relationships and color harmonies into a visual narrative that resonates deeply with human perception. 

While humans can easily perceive and appreciate the beauty of a well-composed image, training neural network models to explicitly understand and control aesthetic composition remains a non-trivial challenge. This challenge stems from two main factors: (1) the subjective nature of aesthetic preferences~\cite{liu2010optimizing}, which lacks a universally accepted quantitative framework for evaluation, and (2) the inherent coupling of composition and semantic content in natural image distributions, making it difficult to model composition independently. In the context of image generation, improving aesthetics often relies on training models with high-quality, visually appealing datasets~\cite{dai2023emu, chen2024pixart, betker2023dalle3}, where aesthetic composition norms are implicitly learned. While this approach yields aesthetically pleasing results, it does not provide explicit control over composition. Additionally, layout-based control methods~\cite{li2023gligen, phung2023grounded, zhou2024migc} can generate compositions that align with user requirements. However, these methods generally require precise spatial constraints related to specific objects, overlooking the fact that composition operates independently of semantic content. In practice, users may not know the exact objects or their arrangements, but instead seek more abstract, high-level control over composition. This necessity underscores the demand for models capable of representing and manipulating composition in a semantic-agnostic manner.

To address this, we propose Composer, a framework designed for controlling composition in image generation. Our framework begins by transferring composition from a reference image through the extraction of composition-aware representations. Based on aesthetic theories and existing aesthetic evaluation studies~\cite{cohen2006color,obrador2010role,liu2010optimizing}, we posit that a good composition involves two key factors: a well-formed \textbf{spatial structure} and a harmonious \textbf{ color distribution}. We designed saliency mapping and clustering operators to separately extract these composition-aware representations in a semantically-agnostic way. Given that spatial structure and color distribution offer complementary information and share spatial properties, we developed a unified, parameter-efficient, and feature-fused conditioning module that enables effective spatial control during generation. Furthermore, we extend composition conditions with additional mask channel, empowering users with customizable editing and region-specific control. To address cases where users specify only text themes without composition reference, we leverage pre-trained Large Vision-Language Models (LVLMs) \cite{wang2024qwen2, liu2024visual} to perform explicit composition planning. Specifically, given a user-specified theme, our framework harnesses the in-context learning capabilities of LVLMs to retrieve reference images that offer insightful compositional cues from an external aesthetic database. Finally, to optimize composition in fully reference‑free scenarios, we perform text-to-composition fine-tuning on the trained control module to learn implicit composition planning, enabling it to infer composition guidance directly from text prompts without requiring explicit reference image.

For model training, we utilized state-of-the-art LVLMs~\cite{wang2024qwen2} and text-to-image models~\cite{blackforest2024} to generate high-quality captions and images for a curated dataset comprising approximately 2 million high-quality, diverse image-text pairs. Experimental results demonstrate that Composer significantly improves the aesthetic quality of generated images. Additionally, Composer supports flexible combination of conditions and the specification of mask regions, enabling more personalized and precise composition control.

\noindent The key contributions of this work are:

\begin{itemize}
    \item We propose a novel framework, Composer, that explicitly models composition representations, enabling composition transfer, planning, and manipulation.
    \item Our framework enables precise composition control by extracting composition-aware representations of spatial structure and color harmony, which are integrated into a unified, parameter-efficient control module.
    \item To support text-to-image generation, we adopt retrieval-based augmentation for explicit composition planning and text-to-composition fine-tuning for implicit composition planning.
    \item We construct a high-quality dataset for training and experimental results demonstrate significant improvements in both aesthetic quality and composition control flexibility in image generation.
\end{itemize}

\begin{figure*}[h!t]
  \centering
   \includegraphics[width=1\linewidth]{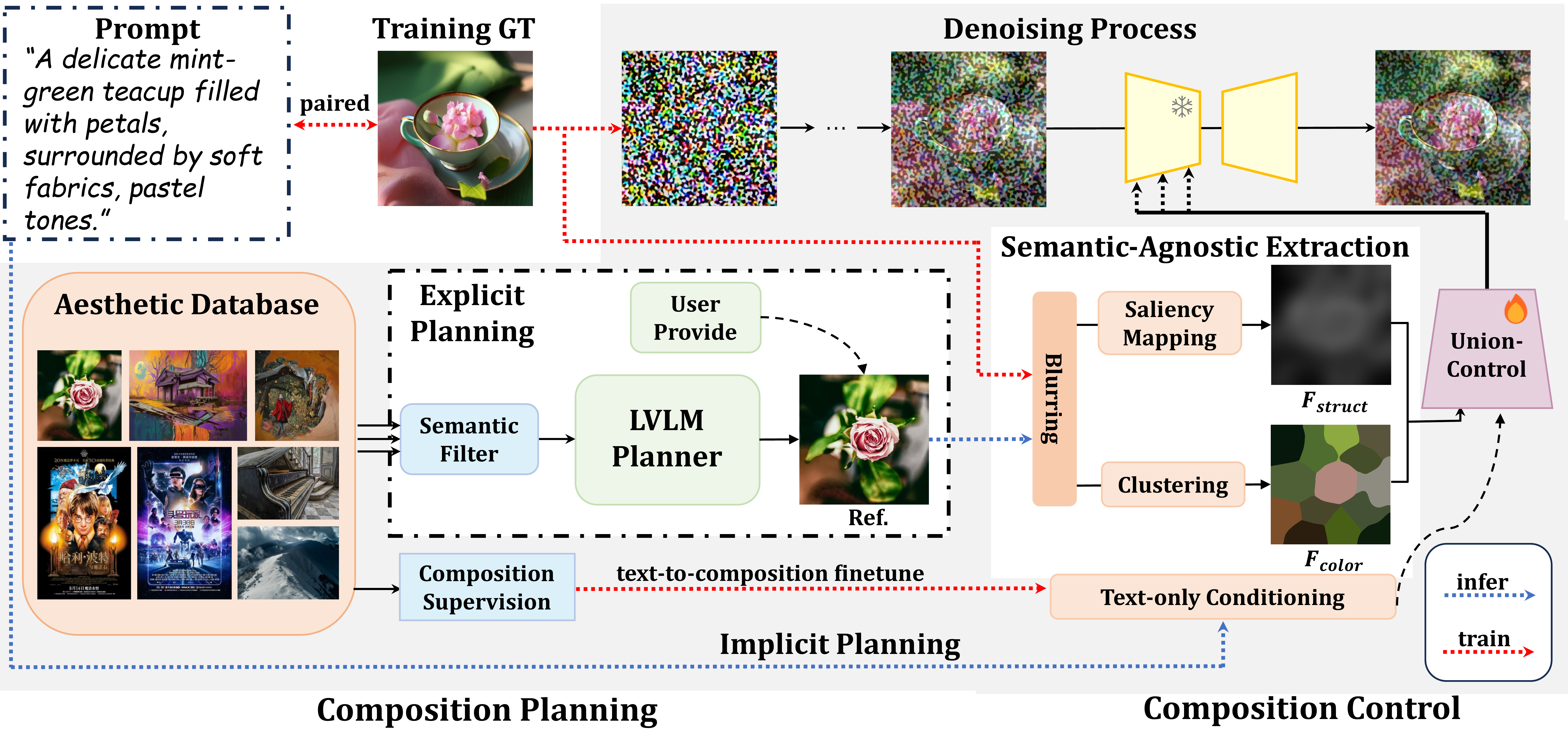}

   \caption{
    Overview of \textbf{Composer} framework, which consists of two main components: 
    (1) \textbf{Composition Control:} Extracting semantic-agnostic composition representations, including spatial structure and color distribution, which are fed into the Union-ControlNet module to generate images that satisfy both the textual and composition conditions.
    (2) \textbf{Composition Planning:} Starting from an external aesthetic database, we obtain a set of candidate images through Semantic Filtering. Given some composition referring examples, we utilize the in-context learning capabilities of Large Vision-Language Models (LVLMs) to perform composition-aware retrieval from the image set based on the given textual theme. Union‑ControlNet is initially trained on paired text–image data to internalize explicit composition conditioning. It is then fine‑tuned with a text‑to‑composition objective to acquire implicit composition planning; during inference, we perform composition planning to achieve text-to-image aesthetic enhancement.
    }

   \label{fig:pipeline}
\end{figure*}
\section{Related Work}
\textbf{Composition in Image Understanding}.
% In the field of image understanding, composition has been extensively studied in the context of image aesthetic assessment (IAA), which can be categorized into deep learning-based and hand-crafted feature-based approaches. Deep learning-based methods, which represent the current state-of-the-art, leverage large datasets and models like CNNs to directly learn aesthetic features from data, achieving superior performance in aesthetic binary classification or scalar scoring tasks [xx][xx]. Advanced models, such as multi-task frameworks [xx] and aesthetic distribution prediction models like NIMA [xx], further capture the nuances of human aesthetic perception. However, these methods primarily model composition implicitly and are not designed for direct manipulation.
Composition has been a key focus in image aesthetic assessment (IAA), explored through both deep learning and hand-crafted feature approaches. Deep learning methods leverage large datasets and neural networks to directly learn aesthetic features, excelling in tasks such as binary classification~\cite{datta2006studying, liu2018deep} and scalar scoring~\cite{kong2016photo, damera2000image}. Advanced models, including multi-task frameworks~\cite{zeng2019unified}, NIMA for aesthetic distribution prediction~\cite{talebi2018nima}, and attention-based architectures~\cite{yang2022maniqa}, capture complex elements of human aesthetic perception but often model composition implicitly, lacking explicit control mechanisms. Hand-crafted methods, grounded in photography and aesthetic theory, define explicit features like 56 visual attributes based on photographic rules~\cite{obrador2010role}. Although these methods are less powerful in representation, they offer clear insights into compositional structures. Inspired by work suggesting spatial composition is guided by the salient region distribution~\cite{liu2010optimizing}, we developed a saliency mapping algorithm to construct semantic-agnostic spatial representations. Additionally, drawing on color harmonization principles~\cite{cohen2006color} and dominant color extraction findings~\cite{he2023thinking}, we incorporate color distribution modeling for enhanced composition control in image generation.

\vspace{2mm}

\noindent\textbf{Composition in Image Generation}. State-of-the-art generative models~\cite{dai2023emu,chen2024pixart,betker2023dalle3} often improve the aesthetic quality of the generated images by constructing post-training datasets with enhanced aesthetics. These datasets are typically obtained by synthesizing or collecting images and then filtering them through aesthetic scoring models~\cite{schuhmann2022laion}. However, this approach is implicit and lacks controllability. In contrast, layout-based control methods ~\cite{li2023gligen, phung2023grounded, zhou2024migc, zhang2025eligen} allow for precise manipulation of object positions, shapes, and interactions. This fine-grained control, however, primarily focuses on object-specific arrangements and overlooks the broader concept of composition, which often operates independently of semantics. While some works~\cite{zhang2024realcompo,yang2024mastering,liu2024draw,liu2025cot} use large language models (LLMs) for layout generation, LLMs are better suited for understanding composition than directly generating it. In addition, several approaches~\cite{chen2025posta, chen2025postercraft} utilize LVLM-based layout planning and text rendering to generate high-quality posters. Nevertheless, such methods are designed for domain-specific graphic design tasks, do not generalize to open-domain image generation, and offer only limited control over global layout. Our approach leverages the in-context learning capabilities of LVLMs to plan compositions through retrieval-based augmentation, facilitating aesthetic enhancement in text-to-image tasks.

\section{Method}
\subsection{Overview}
Our goal is to generate images that align with both the user-provided textual theme and the composition condition. Composer achieves this through a process of composition planning and composition control. As illustrated in Fig.~\ref{fig:pipeline}, given a textual theme, Composer either uses an LVLMs to plan the composition or relies on a user-provided reference image \( I_{\text{ref}} \). Through a Semantic-Agnostic Extraction process, we extract composition-aware representation, specifically the spatial structure \( F_{\text{struct}} \) and color distribution \( F_{\text{color}} \), from the reference image. These features, along with an optional mask \( M \), are fed into the conditioning module, Union-ControlNet, which guides the denoising process. In addition to reference-based generation, Composer supports a reference-free mode in which Union-ControlNet performs implicit composition planning to infer composition guidance directly from textual input.

\subsection{Condition Extraction}

In order to transfer only the composition rather than other elements, it is essential to extract composition representations from the reference image that are independent of semantic content. Based on aesthetic theory, we summarize the subjective principles and rules of composition into two fundamental, quantifiable representations and model them:

\textbf{Spatial Structure:} Good composition adheres to heuristic principles such as the rule of thirds, diagonal composition, and negative space \cite{obrador2010role}. Inspired by previous aesthetic evaluation studies~\cite{liu2010optimizing}, which analyze composition using saliency detection, we assert that these principles guide the distribution and density of primary objects in an image, which we define as spatial structure. Following~\cite{achanta2009frequency}, given a reference image \( I_{\text{ref}} \in \mathbb{R}^{W \times H \times 3} \) in the Lab color space\cite{standard2007colorimetry}, the structural representation is calculated as follows:
\begin{equation}
F(x, y) = \left\| I_{\mu}(x, y) - \left( G_{\sigma, k} * I_{\text{ref}} \right)(x, y) \right\|,
\end{equation}
where \( I_{\mu} = \left( E_{I_{\text{ref}}}[L], E_{I_{\text{ref}}}[A], E_{I_{\text{ref}}}[B] \right)\), \( I_{\text{ref}}(x, y) \) is the reference image at position \((x, y)\), \( * \) denotes the convolution operation, \(\sigma\) is the standard deviation controlling the blur intensity, \( G \) is a Gaussian kernel with the kernel size \(k \times k\). After global normalization, the structural representation \( F_{\text{struct}} \) is obtained. However, such features may encode too much semantic and high-frequency information. To address this, we sample kernel sizes \(k\) over a larger numerical range. This dynamic kernel introduces flexibility and improves generalization, which is beneficial for subsequent manipulations. To stably represent the density distribution while avoiding the influence of large-scale smooth outliers, we localize the expectation operator to match the size of the Gaussian kernel. Additionally, we omit normalization to preserve absolute values across all image distributions rather than in single image. The structural representation $F_{\text{struct}}$ is thus computed as:

\begin{equation}
F_{\text{struct}}(x, y) = \left\| I_{\mu(k)}(x, y) - \left( G_{\sigma, k} * I_{\text{ref}} \right)(x, y) \right\|,
\end{equation}
where \( I_{\mu(k)}(x, y) \) denotes the mean computed locally using the same kernel size \( k \) as the Gaussian filter.

\textbf{Color Distribution:} Color harmony, which refers to the set of colors that have a special internal relationship, provides a pleasant visual perception. As noted in~\cite{cohen2006color, he2023thinking}, it is determined by regional dominant colors and their positional relationships, which we define as color distribution. Linear Iterative Clustering (SLIC)~\cite{achanta2012slic}, which is a superpixel segmentation algorithm that clusters pixels based on both spatial and color similarities, aligns well with the characteristics of regional dominant colors. To filter out semantic information, we apply a larger Gaussian blur before performing superpixel clustering, with both the Gaussian kernel size $k$ and clustering parameters $\theta$ being dynamic hyperparameters:
\begin{equation}
F_{\text{color}} = \text{SLIC}\left( G_{\sigma, k}(I_{\text{ref}}) , \theta\right).
\end{equation}
Thus we obtain a robust representation of the spatial structure and color distribution that is independent of semantic content.

\textbf{Composition as a low-information, semantic-agnostic prior.} A key motivation behind our condition extraction is that complex visual scenes are often supported by simple geometric skeletons. Consequently, we model \emph{composition} as a low-information, semantic-agnostic prior, where coarse spatial structure and global color distribution capture the dominant layout patterns while suppressing fine object-specific details. Under this view, a reference image is not required to enumerate or match the target semantics; instead, it serves as a representative instance of a \emph{composition mode} shared by many different subjects and abstraction levels. This design choice motivates our use of Gaussian blurring and clustering-based aggregation to construct compact, reusable composition conditions that remain effective even when semantically exact references are unavailable.

\subsection{Composition Control}
\label{sec:composition_control}
Spatial structure and color distribution are complementary in terms of information and share spatial characteristics, so we adopt the ControlNet paradigm~\cite{zhang2023adding}, widely used for spatial conditioning. A straightforward implementation would be to employ two parallel ControlNet modules to control the two composition features independently. However, this method has two main issues: it leads to inefficient parameterization, as the parameters of the parallel modules are comparable to those of the original U-Net despite the lower complexity of the composition subspace; and insufficient information utilization, as treating spatial structure and color distribution separately limits their interaction, despite their inherent spatial correspondences.

\begin{figure}[h!t]
  \centering
   \includegraphics[width=.8\linewidth]{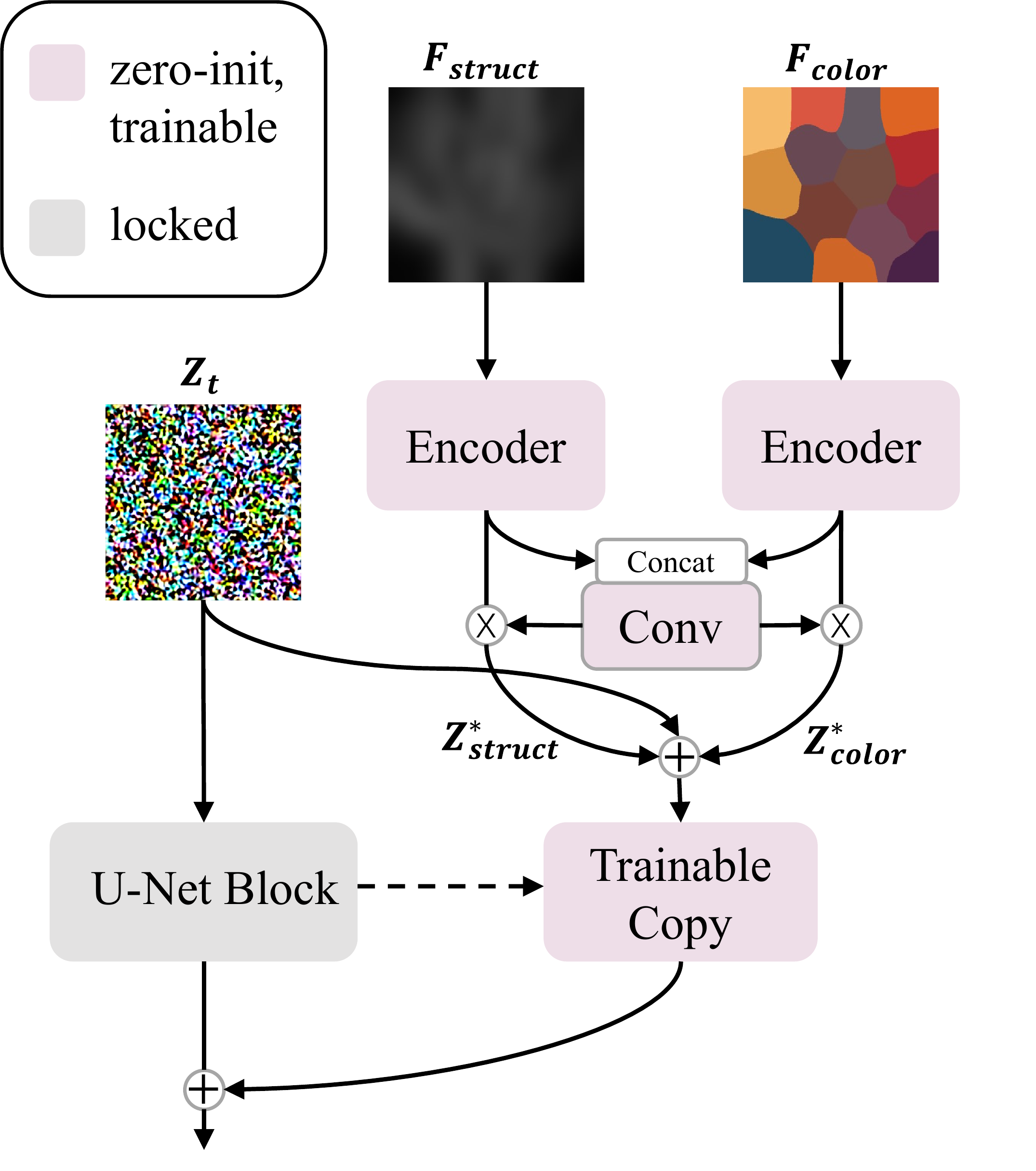}

   \caption{\textbf{Union-ControlNet module.}}
   \label{fig:uni}
\end{figure}
To address these issues, we propose to merge potentially redundant parameters in the aforementioned dual structure. Noting that ControlNet consists of an Encoder and a UNet, where the latter utilizes extensive parameters to align condition features with the latent space of the original UNet, we identify this component as suitable for merging. However, merging the Encoder would compromise the ability to support conditional re-weighting, which is essential for fine-grained control and handling conditional conflicts. To this end, we propose a simple yet efficient module, Union-ControlNet. As illustrated in Fig.~\ref{fig:uni}, the input composition conditions $F_{\text{struct}}$ and $F_{\text{color}}$ are first encoded symmetrically through convolutional encoders into latent representations $Z_{\text{struct}}$ and $Z_{\text{color}}$, aligning their shapes with the noised latent $Z_t$. To enable condition re-weighting and information interaction, we introduce a modulation mechanism to perform rescaling on the latent condition representations. The modulation signal is based on concatenated information from both \( Z_{\text{struct}} \) and \( Z_{\text{color}} \) along the channel dimension with re-weighting coefficient \( w_{\text{struct}} \) and \( w_{\text{color}} \) as follows:

\begin{align}
Z_{\text{struct}}^* 
    &= Z_{\text{struct}} \times \text{ConvModule}([Z_{\text{struct}}, Z_{\text{color}}]) \nonumber\\
    &\quad \times w_{\text{struct}}.
\label{eq:modulation}
\end{align}

A similar process is applied to \( Z_{\text{color}} \). Finally, the modulated latent representations are passed through a unified trainable encoder to introduce composition guidance. Additionally, we extend both \( F_{\text{struct}} \) and \( F_{\text{color}} \) by an extra dimension to accommodate masks, allowing for region-specific conditioning. We will compare this with the parallel (Dual) variant in the experiments to demonstrate the effectiveness of the Union-ControlNet module.

\subsection{Composition Planning}
The goal of composition planning is to derive composition conditions for the given textual prompt \( P_{theme} \), enabling aesthetic enhancement in the text-to-image generation process. Directly training a model to explicitly generate an aesthetically pleasing composition solely from text is highly challenging. However, vast collections of real-world images—such as photographs, artworks, and graphic designs—encapsulate rich compositional knowledge that can be harnessed to guide the composition structure of the target theme.

\noindent\textbf{Explicit Composition Planning}. Given a collection of images \( \mathcal{K} = \{ I_1, I_2, \dots \} \), we utilize a captioning model to associate each image \( I_i \) with a corresponding theme \( P_i \), thereby constructing a set of image-text pairs \( \{(I_1, P_1), (I_2, P_2), \dots \} \). Since LVLMs can understand both the semantic content and the composition structure of images, we employ them to retrieve suitable composition references from the image set \( \mathcal{K} \) based on the target theme \( P_{theme} \). Due to the limited context length of LVLMs, we initially employ CLIP model~\cite{radford2021learning} to filter out images from the original collection \( \mathcal{K} \) that have low semantic relevance to \( P_{theme} \). This step yields a refined subset \( \mathcal{K}_{\text{f}} \), containing candidates that are more closely aligned with the target theme. However, retrieving references solely based on semantic similarity,  often leads to a loss of diversity in composition. To address this, we aim to guide the retrieval process toward a composition-aware perspective, selecting reference images that provide meaningful composition cues beyond mere content alignment. Specifically, the retrieval process is enhanced using Chain-of-Thought (CoT)~\cite{wei2022chain} reasoning. We carefully select composition reference examples presented as \((T_{\text{theme}}, I_{\text{ref}}, I_{\text{res}})\), where \(T_{\text{theme}}\) represents the textual theme, \(I_{\text{ref}}\) is the reference image, and \(I_{\text{res}}\) is the resulting image, as shown in Fig.~\ref{fig:pipeline}. The instruction \(P_{\text{ins}}\) guides the model to reference these examples to facilitate composition-aware retrieval based on \(P_{\text{theme}}\) from \(\mathcal{K}_{\text{f}}\). For LVLMs that can only output text, we prompt them to generate textual output that corresponds to \(I_{\text{ref}}\), as follows: 

\begin{equation}
I_{\text{ref}} = \arg\max_{(I_i, P_i) \in \mathcal{K}_{\text{f}}} \Pr(P_i \mid P_{\text{ins}}, P_{\text{theme}}, \mathcal{D}, \mathcal{K}_{\text{f}}),
\end{equation}
where \(\mathcal{D} = \{(T_{\text{theme}}, I_{\text{ref}}, I_{\text{res}})_j \mid j \leq m \}\) is the set of triplet examples.

\begin{figure}[h!t]
  \centering
   \includegraphics[width=1.0\linewidth]{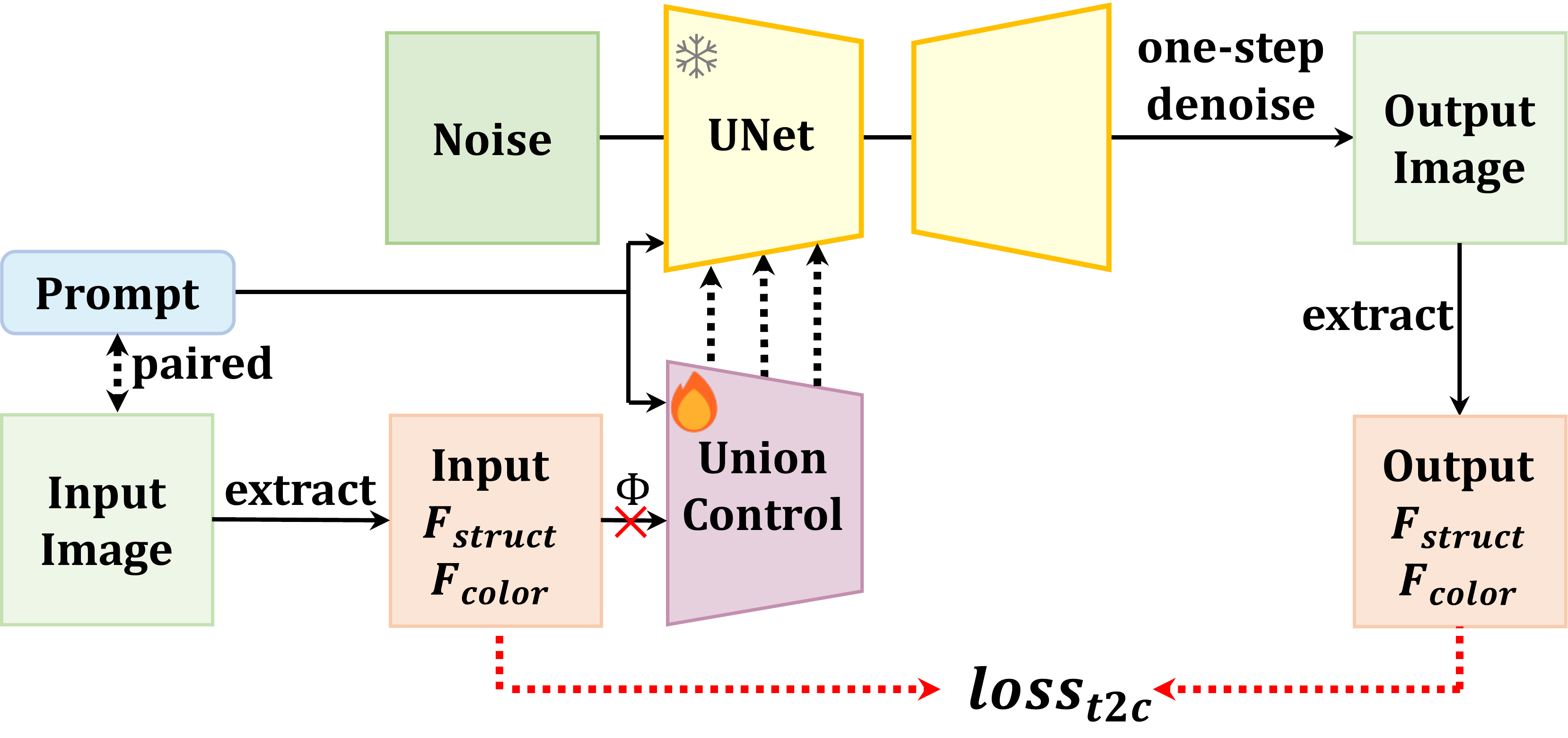}

   \caption{\textbf{text‑to‑composition fine‑tuning for implicit composition planning.}}
   \label{fig:t2c}
\end{figure}

\noindent\textbf{Implicit Composition Planning}. The trained Union‑ControlNet preserves the UNet‑style cross‑attention layers necessary to accept textual conditioning. Because its learned representations are already specialized for composition control, we can further fine‑tune the network under \emph{text‑only} inputs to establish a direct text‑to‑composition mapping. As illustrated in Fig.~\ref{fig:t2c}, we supervise this adaptation with the loss

\begin{align}
\mathcal{L}_{\text{t2c}}
  &= \operatorname{dis}\bigl(
       F_{\text{struct}}(x_0),\, F_{\text{struct}}(x_0')
     \bigr) \nonumber\\[2pt]
  &\quad+\;
     \operatorname{dis}\bigl(
       F_{\text{color}}(x_0),\, F_{\text{color}}(x_0')
     \bigr).
\label{eq:t2c_loss}
\end{align}

where $x_0$ denotes the ground‑truth clean image and $x_0'$ is the one‑step denoised prediction generated by the model $\epsilon_{\theta}$ at timestep $t$ with text condition $c_{\text{text}}$ and the composition token $\phi$ omitted:

\begin{equation}
x_0'
  \;=\;
  \frac{
    x_t' \;-\; \sqrt{1 - \alpha_t}\,
    \epsilon_\theta\!\bigl(x_t',\, \phi,\, c_{\text{text}},\, t\bigr)
  }{
    \sqrt{\alpha_t}
  }.
\end{equation}
Here, $\operatorname{dis}(\cdot,\cdot)$ is a feature‑space distance metric (e.g., $\ell_2$ norm), while $F_{\text{struct}}$ and $F_{\text{color}}$ extract structural and chromatic descriptors, respectively. This objective encourages the model to infer composition-aware guidance solely from the textual prompt.

\subsection{Training}
\begin{table*}[ht]
    \centering
    \begin{tabular}{lccc}
    \toprule
    Source & Samples & Prompts & Images \\
    \midrule
    LLaVA-Next finetune~\cite{liu2024llava} & ~800K & Re-captioned using Qwen2-VL & Original images \\
    LLaVA-pretrain~\cite{liu2024visual} & ~600K & Original prompts &  generated by Flux-dev~\cite{blackforest2024} \\
    ProGamerGov synthetic~\cite{proGamerGov2023synthetic} & ~600K & Re-captioned using Qwen2-VL & Filtered for validity \\
    \midrule
    Overall & ~2M & Mixed & Mixed \\
    \bottomrule
    \end{tabular}
    \caption{Composition of the curated training dataset.}
    \label{tab:dataset}
\end{table*}
\noindent\textbf{Datasets}. Considering the parameter scale of Union-ControlNet and the task complexity, we curated and processed approximately 2 million image-text pairs to construct a diverse and high-quality dataset. The dataset is composed of images and captions from several sources, including LLaVA-pretrain~\cite{liu2024visual}, LLaVA-next fine-tuning~\cite{liu2024llava}, and ProGamerGov synthetic~\cite{proGamerGov2023synthetic}. We respectively used Qwen2-VL~\cite{wang2024qwen2} and Flux1.0-dev~\cite{blackforest2024} to enhance the quality of the caption and images. Details of the dataset composition are provided in Tab.~\ref{tab:dataset}.

\noindent\textbf{Training Strategy.} As shown in Fig~\ref{fig:pipeline}, during training, the ground truth (GT) images paired with text are input into the Semantic-Agnostic Extraction process. The resulting composition conditions, along with the text, jointly guide the model to perform denoising. The model is trained using a standard diffusion loss, following~\cite{zhang2023adding}. Since the learning difficulty differs between spatial structure and color distribution, we adopt a two-stage curriculum to accelerate convergence. In the first stage, we randomly train each condition with the others masked, according to the training steps. Once the loss stabilizes, we proceed to the second stage, where we progressively increase the probability of providing all conditions simultaneously until joint conditioning is used 100\% of the time. After this supervised phase, we further perform text‑to‑composition fine‑tuning to endow the model with implicit composition‑planning capability, enabling it to derive compositional guidance directly from textual prompts.

\subsection{Applications}

\begin{figure}[h!t]
  \centering
   \includegraphics[width=1\linewidth]{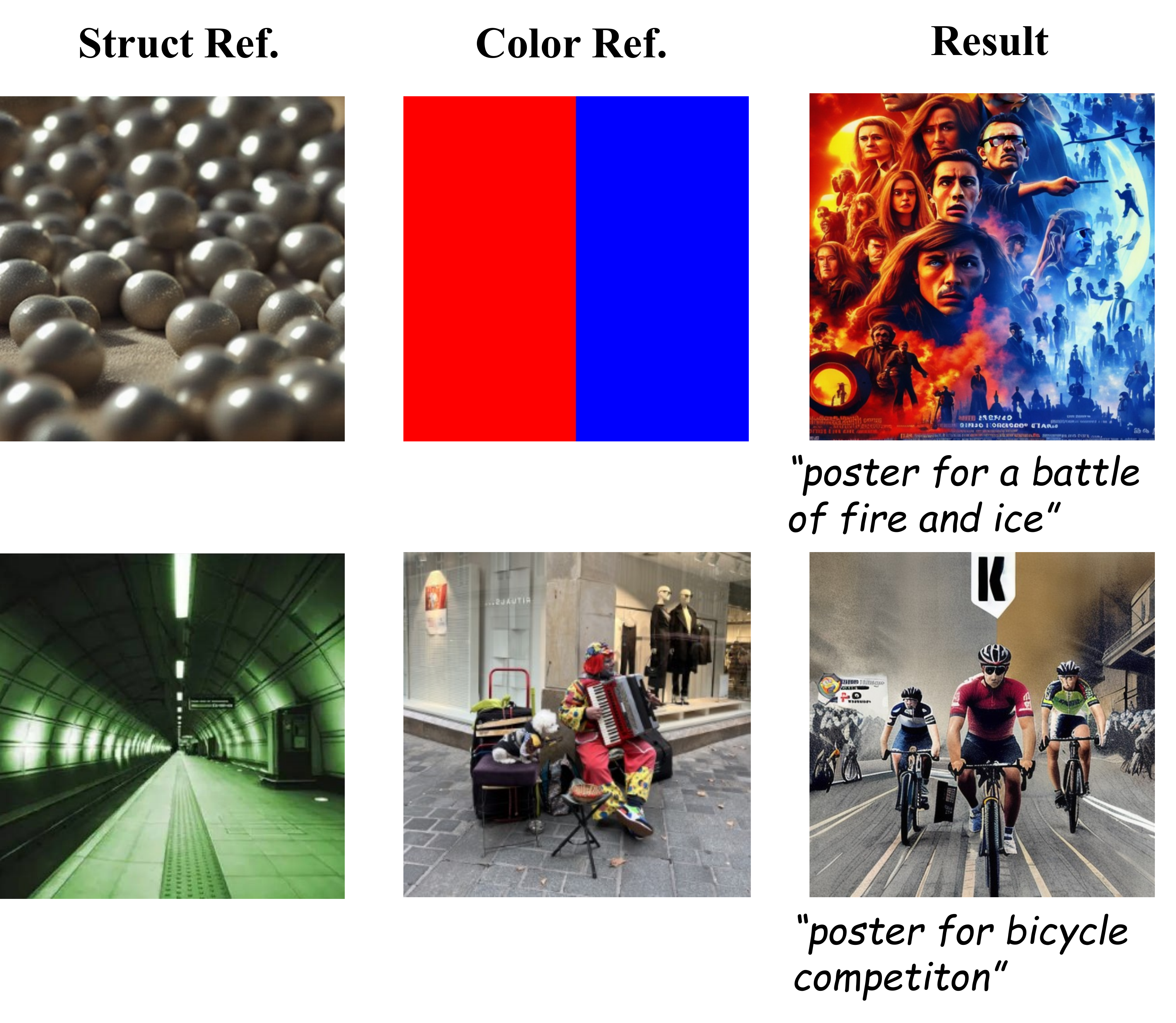}

   \caption{\textbf{Combined Composition Transfer.} The generated result is guided by the spatial structure from one reference image (\textbf{Struct. Ref.}) and the color distribution from another reference image (\textbf{Color Ref.}), demonstrating the ability to integrate different composition elements.
}
   \label{fig:2img}
\end{figure}

\begin{figure}[h!t]
  \centering
   \includegraphics[width=1\linewidth]{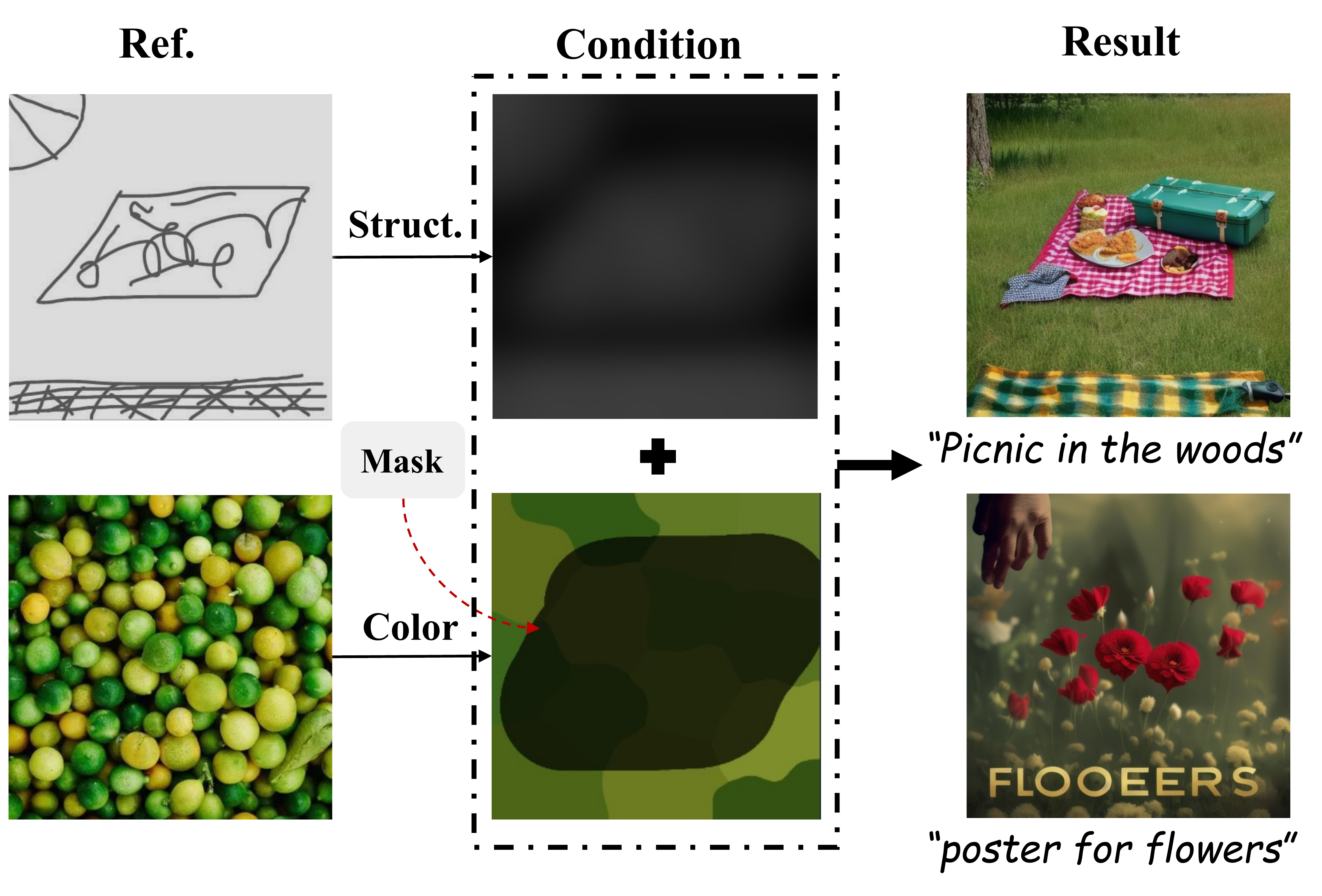}

   \caption{    \textbf{Composition Manipulation.} From left to right: the reference image (\textbf{Ref.}), the extracted composition representation (\textbf{Condition}), and the generated results under different textual themes with combined composition control.
}
   \label{fig:edit}
\end{figure}

\noindent\textbf{Composition Transfer}. Composer enables transferring composition from reference images. For a single reference image (Fig.~\ref{fig:1img}), Composer can generate high-quality images under diverse text prompts and varying composition conditions. Furthermore, as demonstrated in Fig.~\ref{fig:2img}, Composer supports combined control, where the spatial structure and color distribution can be extracted from separate reference images.

\noindent\textbf{Composition Manipulation}. Composer also facilitates more flexible control methods, such as extracting composition from user-drawn sketches or modifying the composition of an existing image using masks. Fig.~\ref{fig:edit} illustrates personalized results obtained by combining different control inputs, showcasing the versatility of the framework in meeting user-specific requirements.

\section{Experiments}
\noindent\textbf{Implementation Details}. For feature extraction of spatial structure and color distribution, considering that the input images are primarily distributed at 512x512, we uniformly adopt a Gaussian kernel size sampled from the range $[193, 321]$ with an interval of 16. For the SLIC algorithm, we empirically set the region size to 128 and the number of iterations to 20. To train the conditional mask based on additional channels, considering that user masks in practical scenarios tend to be continuous, we draw a randomly sized ellipse or rectangle on the image space and randomly invert it as the mask input. We implement composition-guided generation based on Latent Diffusion Models (LDM)~\cite{rombach2022highresolution} as the backbone. The training process adopts a fixed learning rate of \(1 \times 10^{-5}\). By default, we employ the DDIM sampler for a 50-step denoising process during inference. For composition planning, we utilize the Qwen2-VL~\cite{wang2024qwen2} model. For composition planning using LVLM, we provide an example prompt of the chain of thought as follows:

\begin{quote}
\small
\noindent\textit{``You are a master of aesthetic composition. I will provide you with a set of candidate composition images and a target theme. Your task is to plan a suitable composition reference image based on the theme. The composition reference follows a process similar to the following: $<T_{theme}, I_{ref}, I_{res}>$. Let's think step by step:}

\noindent\textit{1. Define candidate compositions and color schemes for the given theme.}

\noindent\textit{2. Plan the most suitable composition reference based on the theme, composition, and color scheme. (This avoids reducing the planning process to simple semantic matching.)''}
\end{quote}

\noindent\textbf{Evaluation}. We assess the model from two key perspectives: (1) \textit{Image Quality}, including aesthetics and diversity, measured using Fréchet Inception Distance (FID)~\cite{heusel2017gans} on MS-COCO. Aesthetic quality is further evaluated using Aesthetic Score (Aes)~\cite{schuhmann2022laion} on PartiPrompts, based on a predictor trained on LAION. (2) \textit{Condition Faithfulness}, evaluated through Human Preference Score v2 (HPS v2)~\cite{wu2023human} and ImageReward~\cite{xu2024imagereward} for text-image alignment, both incorporating human aesthetic preferences. Cycle consistency loss~\cite{li2025controlnet} is used to assess adherence to composition conditions in the ablation study. We evaluate our method on two datasets: MS-COCO~\cite{lin2014microsoft}, a large-scale benchmark for object detection and captioning, and PartiPrompts~\cite{yuscaling}, a text-to-image dataset with 1632 prompts and detailed critiques. We also conduct human evaluations to validate image quality and semantic accuracy.

\noindent\textbf{Compared Methods}. We adopt Stable Diffusion 1.5 (SD-v15) as the baseline model. To verify the impact of the dataset, we further fine-tune SD-v15 with LoRA~\cite{hu2021lora} under the same training configuration, denoted as \textbf{SD-v15-ft}. Additionally, we compare against layout-based generation methods, which offer more specific and semantically guided control than composition. These include \textbf{RealCompo}~\cite{zhang2024realcompo} and \textbf{RPG}~\cite{yang2024mastering}. RealCompo utilizes a large language model for layout planning, followed by a layout-to-image model for spatial control. The generated layout is combined with text-to-image outputs to enhance both realism and aesthetics. RPG iteratively generates layouts using LVLMs. This involves a recaptioning process and layout planning, enabling reasonable and semantically coherent compositions. We also evaluate our method against reference-based guidance models, \textbf{IPAdapter}~\cite{ye2023ip}, which injects information from a reference image via cross-attention mechanisms. To reduce the model’s dependence on semantic information to adapt to the setting of our task, we empirically set the guidance weight \(w = 0.5\) for IPAdapter. For all methods mentioned above, the LVLM used for layout and planning is Qwen2-VL~\cite{wang2024qwen2}, while the base generation model remains SD-v15~\cite{rombach2022highresolution} to ensure consistent evaluation.

\begin{table*}[ht]
\centering
\begin{tabular}{lcccccc}
\toprule
\multirow{2}{*}{\textbf{Method}} & \multicolumn{3}{c}{\textbf{MS-COCO}~\cite{lin2014microsoft}} & \multicolumn{3}{c}{\textbf{Parti}~\cite{yuscaling}} \\ 
\cmidrule(r){2-4} \cmidrule(r){5-7}
 & \textbf{CLIP Score} ↑ & \textbf{FID}↓ & \textbf{Aes.} ↑ & \textbf{HPS v2}↑ & \textbf{ImageReward}↑ & \textbf{Aes.} ↑ \\ 
\midrule
SD-v15~\cite{rombach2022highresolution}       & 0.271 & 9.93  & 5.43 & 0.279 & 0.52 & 5.32 \\
SD-v15-ft    & 0.273 & 9.88 & 5.52 & 0.281 & 0.77 & 5.39 \\
RealCompo~\cite{zhang2024realcompo}   & 0.279 & 13.57 & 5.72 & 0.291 & 0.89 & 5.63 \\
RPG~\cite{yang2024mastering}        & \textbf{0.284} & 12.79 & 5.81 & 0.287 & \textbf{1.12} & 5.84 \\
IPAdapter~\cite{ye2023ip}  & 0.270 & 15.82 & 5.66 & 0.281 & 0.29 & 5.52 \\
Composer    & 0.281 & 11.04 & \textbf{5.89} & \textbf{0.295} & 1.09 & \textbf{5.93} \\
Composer-ref. free    & 0.283 & \textbf{9.72} & 5.82 & 0.293 & 1.11 & 5.88 \\
Composer-XL   & 0.298 & 10.25 & 6.02 & 0.296 & 1.17 & 5.99 \\
\bottomrule
\end{tabular}
\caption{\textbf{Quantitative Comparison Results.} \emph{Composer} denotes our framework with \textbf{explicit} composition planning, while \textbf{Composer-reference free} denotes the \textbf{implicit} (reference‑free) variant. We report text-image alignment (\textbf{CLIP Score}~\cite{radford2021learning}, \textbf{HPS v2}~\cite{wu2023human}, \textbf{ImageReward}~\cite{xu2024imagereward}), image quality (\textbf{FID}~\cite{heusel2017gans}), and aesthetic appeal (\textbf{Aesthetic Score}~\cite{schuhmann2022laion}, \textbf{HPS v2}~\cite{wu2023human}, \textbf{ImageReward}~\cite{xu2024imagereward}). Notably, IPAdapter uses the same composition planner of our framework to retrieve reference images.}
\label{tab:comparison}
\end{table*}

\begin{figure*}[h!t]
  \centering
   \includegraphics[width=1\linewidth]{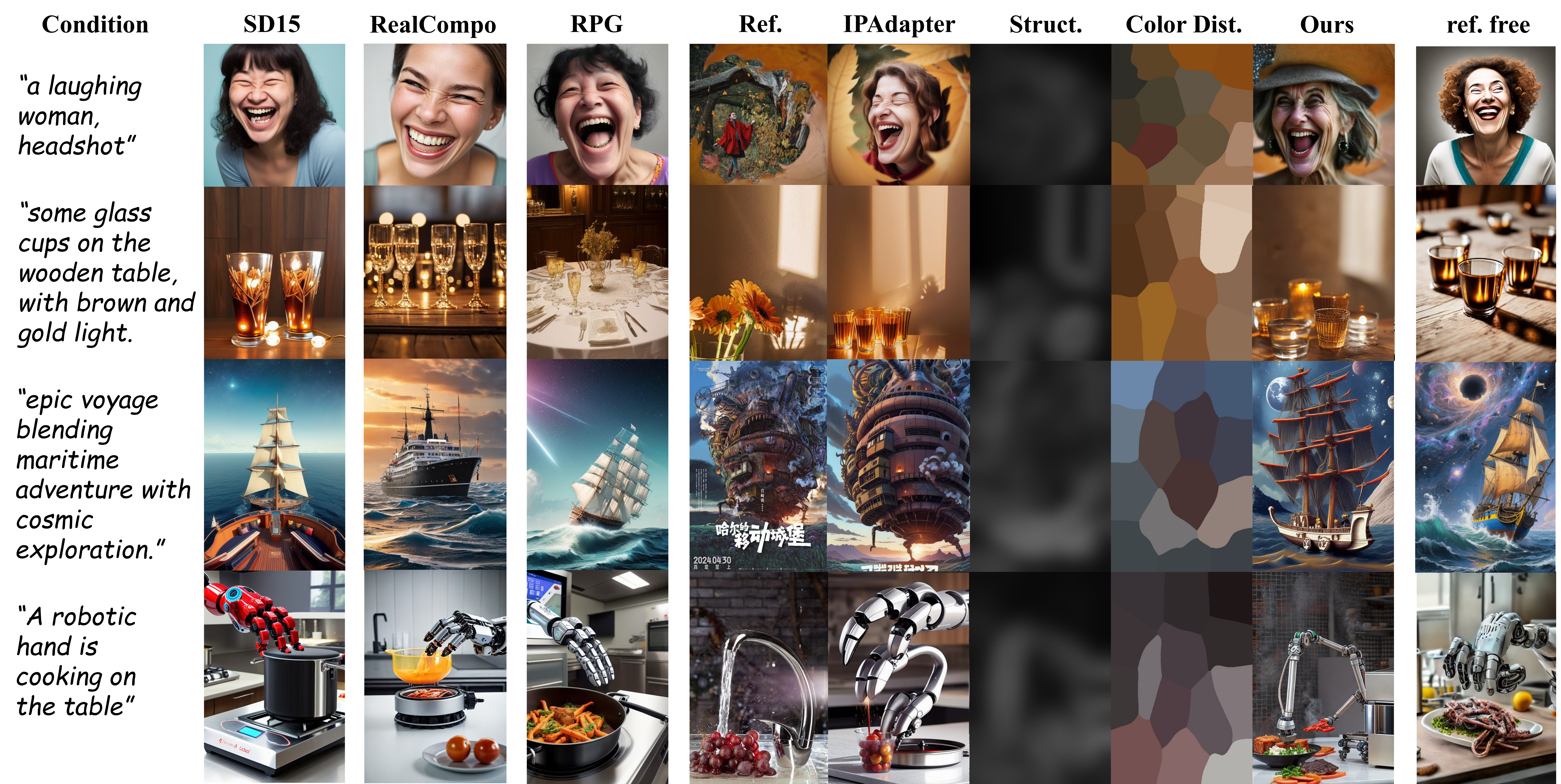}

   \caption{    \textbf{Qualitative Comparison between Composer and Baseline Methods.} \emph{Composer} denotes our framework with \textbf{explicit} composition planning, while \emph{ref. free} denotes the \textbf{implicit} (reference‑free) variant. Both IPAdapter and our method use the same reference image (\textbf{Ref.}), with the corresponding composition representations denoted as \textbf{Struct.} (spatial structure) and \textbf{Color Dist.} (color distribution). 
}
   \label{fig:quality}
\end{figure*}

\subsection{Quantitative Results}

Table~\ref{tab:comparison} presents the evaluation results of the proposed method and several baselines on the MS-COCO and Parti datasets. Specifically, on the MS-COCO dataset, Composer significantly improves both text-image alignment and the aesthetic appeal compared to SD15. Moreover, it outperforms SD15-ft by \textbf{+6.2\%} and IPAdapter by \textbf{+4\%} in aesthetic quality (\textbf{Aes}), demonstrating the effectiveness of our composition learning approach. Composer also achieves competitive results in the CLIP Score compared to RealCompo and RPG, two baseline methods that employ instance-level fine-grained layout control. Additionally, Composer incurs only a minimal penalty in FID compared to other control-based methods, indicating its ability to maintain generalization to various themes without degrading visual quality. On the Parti dataset, Composer achieves the highest scores in both \textbf{HPS v2} and \textbf{ImageReward}, outperforming SD15-ft and IPAdapter. Composer even surpasses layout-control methods such as RPG on these metrics, suggesting that our approach effectively improves the aesthetic appeal of generated images in text-to-image generation. Composer additionally attains the highest \textbf{Aes} score across all methods, highlighting its capability to generate aesthetically superior images aligned with the given text descriptions. Importantly, the reference‑free variant, \textbf{Composer-reference free}, attains the best FID—exceeding SD‑v1.5—demonstrating strong diversity and perceptual quality attributed to the absence of explicit spatial constraint. Its performance on HPS‑v2 and CLIP Score remains comparable to the explicit composition planning version, with only a marginal gap in Aesthetic Score, despite the latter benefiting from access to an external aesthetic database.

\subsection{Qualitative results}

\begin{figure}[ht]
    \centering
    \includegraphics[width=1\linewidth]{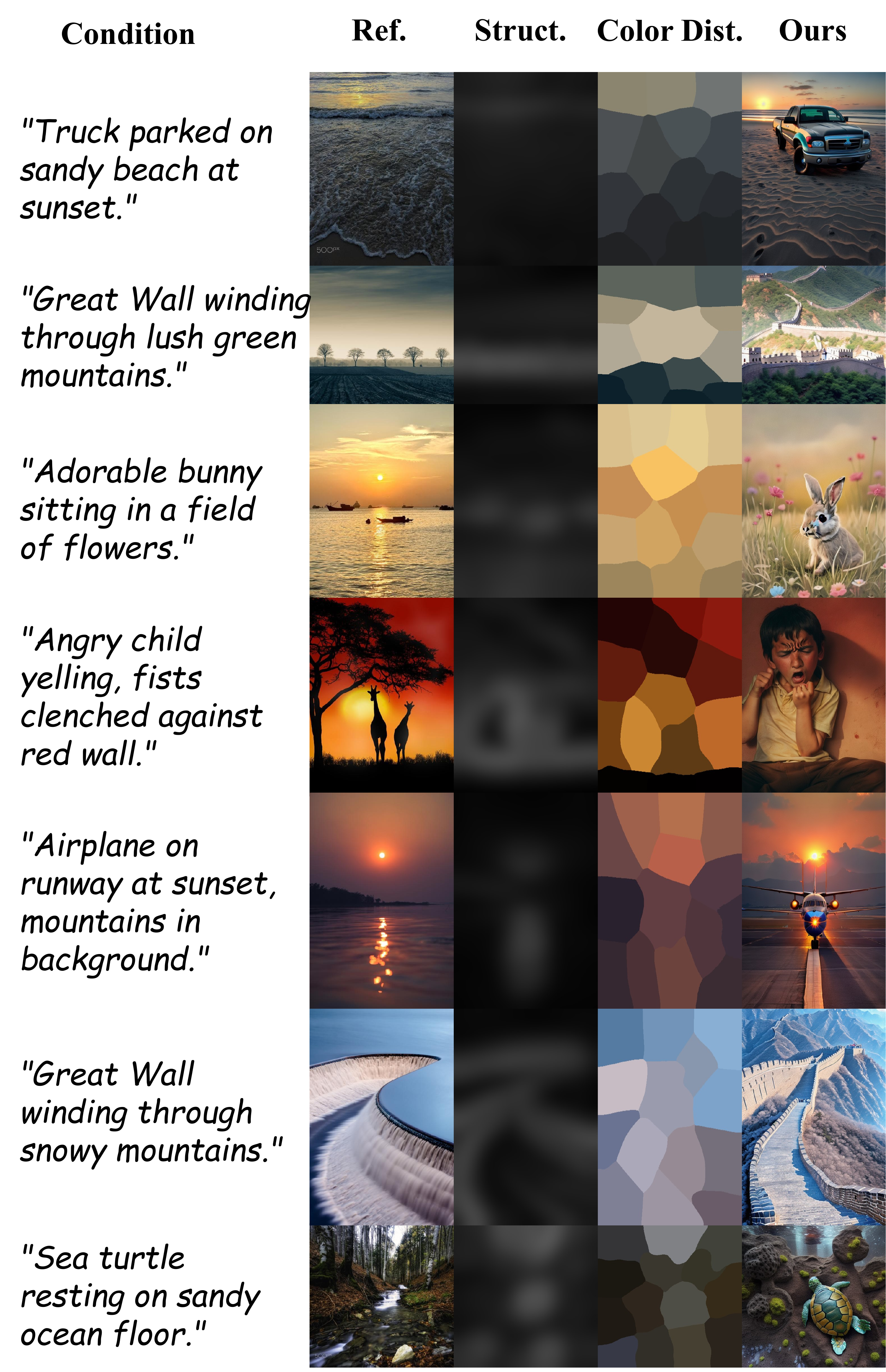}
    \caption{\textbf{More Results of composition transfer.}}
    \label{fig:more1}
    % \vspace{-10pt}
\end{figure}

\begin{figure}[ht]
    \centering
    \includegraphics[width=1\linewidth]{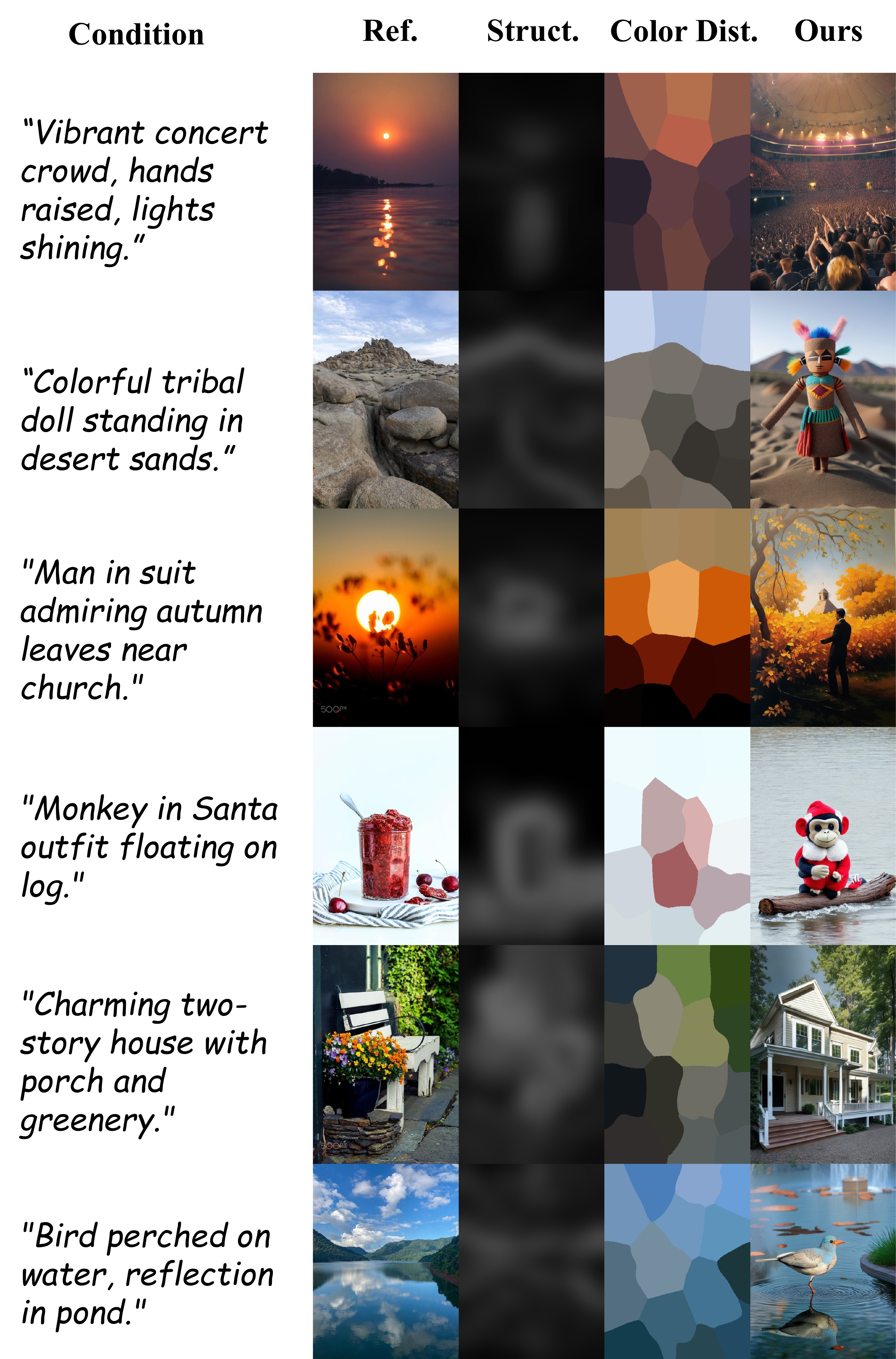}
    \caption{\textbf{More Results of composition transfer.}}
    \label{fig:more2}
    % \vspace{-10pt}
\end{figure}

Fig.~\ref{fig:quality} presents qualitative visual comparisons of the results generated by our framework, \textit{Composer}, alongside several baseline methods. For each generated image, we also display the corresponding reference composition (Ref.), spatial structure map (Struct.), and color distribution map (Color Dist.) used as guidance conditions. In comparison with SD15, RealCompo, and RPG, our method exhibits more visually appealing compositions that align with the input textual descriptions. Composer effectively integrates composition control into the generation process, producing images with both improved spatial coherence and enhanced aesthetic appeal. In contrast with IPAdapter, Composer relies more on reference-based composition planning rather than semantic replication. For example, IPAdapter directly reflects "leaves" and "grapes" from the reference in the first and fifth rows, showing a semantic referring. In contrast, Composer balances composition guidance and semantic relevance, aligning with both the reference and the given theme. For explicit composition planning, we further provide additional visualization results to demonstrate the effectiveness of our method, as shown in Figures~\ref{fig:more1} and~\ref{fig:more2}.

\begin{figure}[ht]
    \centering

    \includegraphics[width=1.0\linewidth]{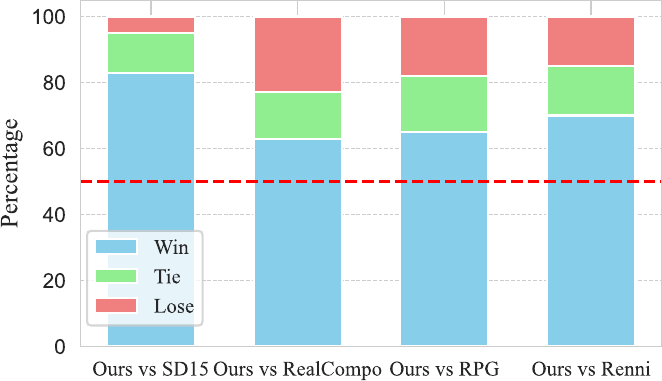}
    \caption{\textbf{User Study} on Text Faithfulness and Aesthetic Quality.}
    \label{fig:human}

\end{figure}

\noindent\textbf{Human Evaluation.} We conducted a comprehensive user study to compare Composer with SD-v1.5 and state-of-the-art layout-to-image methods\cite{zhang2024realcompo, yang2024mastering, feng2024ranni}. For each pair of images generated by two random models, human evaluators were asked to indicate their preference based on both \textit{Text Faithfulness} and \textit{Aesthetic Quality}. A total of 35 participants took part in the study, providing 3,360 votes across 24 diverse prompts. As shown in Fig.~\ref{fig:human}, Composer demonstrates a clear advantage in human preference over the competing methods.

\begin{table*}[ht]
\centering
\caption{\textbf{Ablation Results.} Comparison of conditioning module architectures and composition planning methods.}
\resizebox{\linewidth}{!}{
\begin{tabular}{lccccccccc}
\toprule
\textbf{\#} &\textbf{Plan} & \textbf{Arc.} & \textbf{$L_{\text{struct}}$} ↓ & \textbf{$L_{\text{color}}$} ↓ & \textbf{CLIP} ↑ & \textbf{HPS v2}~\cite{wu2023human} ↑ & \textbf{I. R.}~\cite{xu2024imagereward} ↑ & \textbf{Aes.}~\cite{schuhmann2022laion} ↑ & \textbf{step/s} ↑ \\ 
\midrule
1 &- & - & 112.7 & 108.5 & 0.273 & 0.281 & 0.77 & 5.39 & \textbf{15.48} \\
2 &w/o LVLM & Union & - & - & 0.278 & 0.288 & 0.91 & 5.82 & - \\
3 &w. LVLM & Dual & \textbf{35.9} & \textbf{87.4} & 0.279 & 0.290 & 0.96 & 5.78 & 4.16 \\
\midrule
4 &w. LVLM & Union & 36.8 & 89.2 & \textbf{0.281} & \textbf{0.295} & \textbf{1.09} & \textbf{5.93} & 11.69 \\
\bottomrule
\end{tabular}
}
\label{tab:ablation}
\end{table*}

\subsection{Ablation study}
\label{sec:ablation}
We conduct ablation studies to assess the impact of the composition planner and control module design. Specifically, we compare the Dual-ControlNet (parallel structure) and Union-ControlNet (fused structure) discussed in Section~\ref{sec:composition_control}. Results in Tab.~\ref{tab:ablation} show that while the Dual structure achieves slightly lower scores for Cycle consistency loss~\cite{li2025controlnet} \textit{L\textsubscript{struct}} and \textit{L\textsubscript{color}}, the Union structure outperforms in CLIP Score, HPS v2, ImageReward, and significantly in Aesthetic Score, indicating enhanced text-image alignment and aesthetic quality. Union-ControlNet also demonstrates superior inference speed, confirming its effectiveness and parameter efficiency. Comparisons of \#1, \#2, and \#4 illustrate that using LVLM for composition planning yields better results than semantic filtering alone.

\begin{figure}[ht]
    \centering
    \includegraphics[width=1\linewidth]{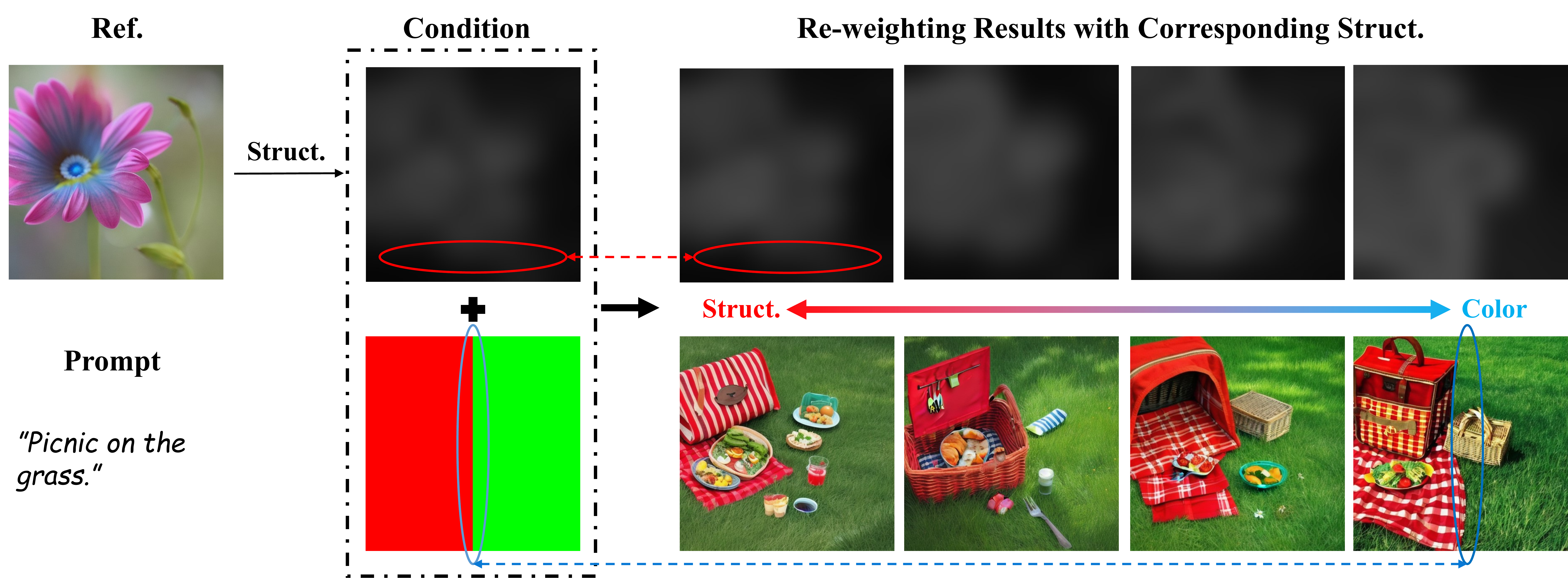}
    \caption{\textbf{Example of Conflicting Conditions:} When structural conditions (e.g., the main subject positioned in the upper-left region with a smooth background) conflict with color conditions (e.g., a sharp gradient transition along the central axis), users can perform customizable condition re-weighting. The red arrows indicate an increased weight for structural conditions, while the blue arrows represent a bias toward color conditions. Note the correspondence between the regions highlighted by the red and blue ellipses.}
    \label{fig:conflict}

\end{figure}

\noindent\textbf{Handling Conflicting Conditions.} To address condition conflicts, our method provides fine-grained control, enabling users to determine the trade-off between conflicting conditions through two mechanisms:

\begin{itemize}
\item \textit{Condition Re-weighting:} The modulation mechanism in Union-ControlNet (Eq.\ref{eq:modulation}) enables users to apply additional re-weighting for conflicting conditions, as shown in Fig. \ref{fig:conflict}. This allows users to prioritize specific conditions during conflicts by fine-tuning their influence.
\item \textit{Regional Control:} By leveraging a condition masking mechanism (shown in the second row of Fig.~\ref{fig:edit}), users can selectively mask out conflicting regions to prevent unintended results.
\end{itemize}

\begin{figure}[h!t]
  \centering

   \includegraphics[width=1\linewidth]{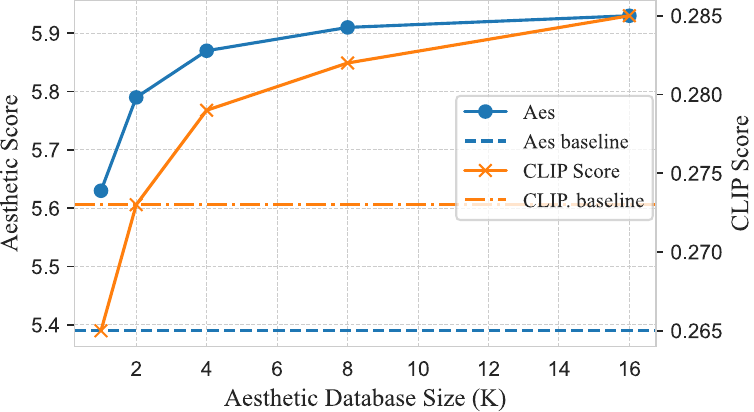}

   \caption{\textbf{Impact of Aesthetic Database Size on Aes. and CLIP Scores.} The baseline is sd-v15-ft. Notably, even a dataset of 16K images has minimal impact on efficiency, as we precompute low-dimensional representations for each image using the CLIP model. Given a specified theme, we filter a fixed subset \(\mathcal{K}_{\text{f}}\), a process that takes less than one second.
}
   \label{fig:size}

\end{figure}

\noindent\textbf{Impact of Dataset Size.} We further analyze the impact of dataset size on image aesthetics and text-image alignment (Fig.~\ref{fig:size}), with SD-v15 as the baseline. When the number of aesthetic images is below 1K, the limited diversity causes generated results to poorly match varied text themes, leading to a low CLIP Score despite an Aesthetic Score higher than the baseline. As the number of images increases, both the CLIP Score and Aesthetic Score improve rapidly, stabilizing after exceeding 10K images. Additionally, due to semantic-based filtering, increasing the number of images has minimal impact on overall processing time.

\noindent\textbf{Runtime Analysis.} We measured the average runtime of Composer with explicit composition planning. The total latency of the full pipeline is 7.7 seconds, with a detailed breakdown provided in Table~\ref{tab:runtime_analysis}. For the reference-free variant, which excludes the LVLM planning stage, the average runtime is reduced to 4.6 seconds.

\begin{table}[ht]
    \centering
    \caption{Runtime Breakdown of the Pipeline (in seconds)}
    \label{tab:runtime_analysis}
    \begin{tabular}{|l|c|}
        \hline
        \textbf{Component} & \textbf{Latency (s)} \\
        \hline
        VLM Planning & 3.1 $\pm$ 0.2 \\
        Structural Operator & $\leq$ 0.2 \\
        Color Operator & $\leq$ 0.2 \\
        Denoising (50 steps) & 4.2 $\pm$ 0.1 \\
        \hline
        \textbf{Total} & 7.7 \\
        \hline
    \end{tabular}
\end{table}

% =========================================================
% Insert this block at the end of Sec. 4 (after \subsection{Ablation study})
% Recommended insertion point: right after the current Table~3 / ablation discussion.
% =========================================================

\subsection{Experiments on SD-XL Backbone}
\label{sec:sdxl}
\begin{figure*}[t]
    \centering
    % Replace with your actual file (pdf/png). Keep consistent column order with caption.
    \includegraphics[width=\linewidth]{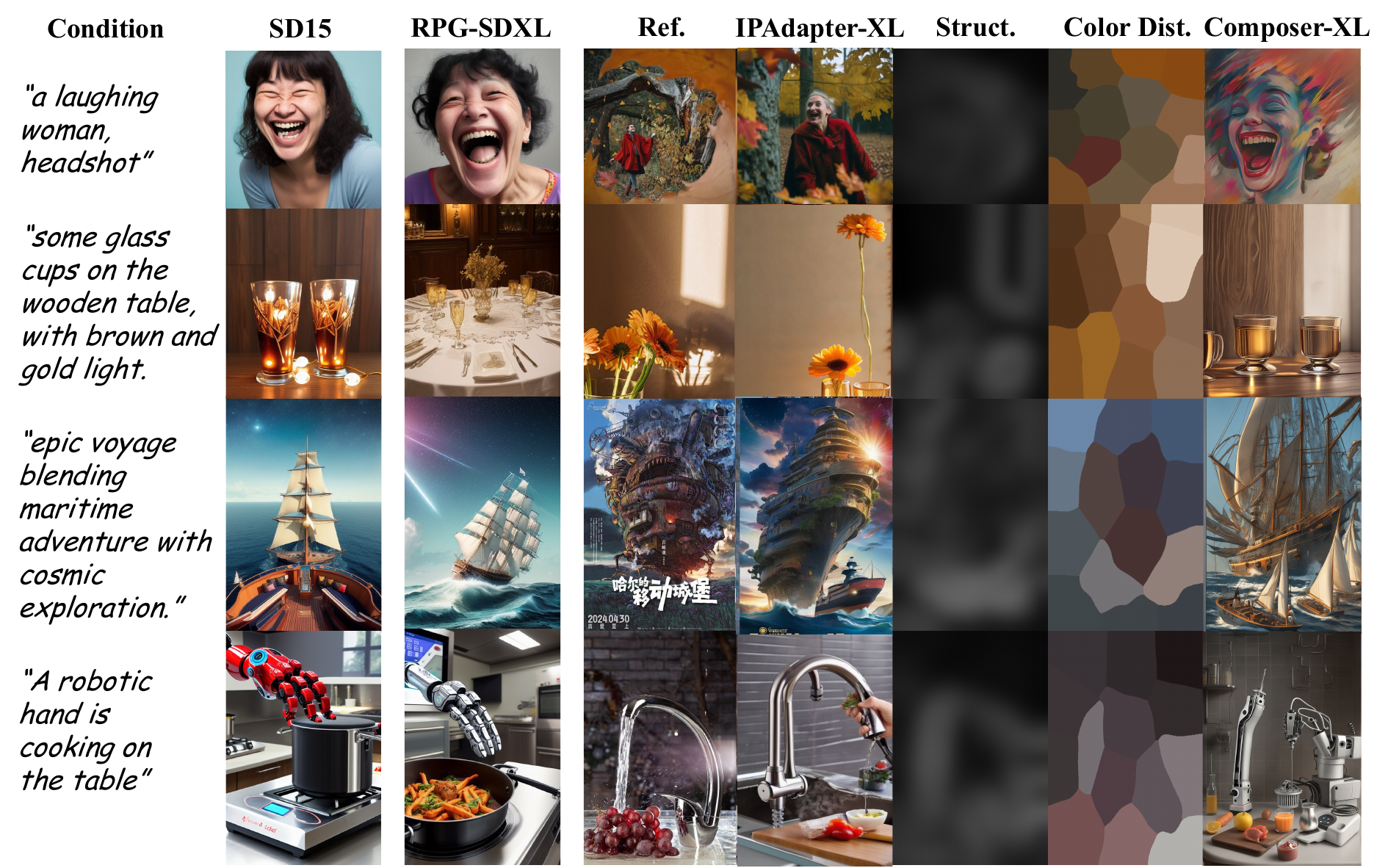}
    \caption{\textbf{Qualitative comparison on SDXL.}
    From left to right: prompt/condition, SD-v1.5 baseline, RPG-SDXL, retrieved reference,
    IP-Adapter-XL, extracted structure, extracted color distribution, and Composer-XL.
    Composer-XL better preserves compositional aesthetics while maintaining strong text consistency.}
    \label{fig:sdxl_qual}
\end{figure*}

\begin{table*}[t]
    \centering
    \caption{\textbf{Conditioning-module ablation on SDXL.}
    Dual-ControlNet improves over the SDXL backbone but is substantially slower;
    Union-ControlNet-XL achieves the best accuracy--efficiency trade-off.}
    \label{tab:sdxl_ablation}
    \setlength{\tabcolsep}{3.8pt}
    \begin{tabular}{lccccccc}
        \toprule
        Method & $L_{\text{struct}}\downarrow$ & $L_{\text{color}}\downarrow$ &
        CLIP $\uparrow$ & HPS v2 $\uparrow$ & ImageReward $\uparrow$ & Aes. $\uparrow$ & step/s $\uparrow$ \\
        \midrule
        SDXL backbone & --   & --   & 0.285 & 0.283 & 0.65 & 5.76 & 3.44 \\
        Dual-ControlNet & 31.2 & 75.9 & 0.294 & 0.295 & 1.12 & 5.91 & 0.90 \\
        \textbf{Union-ControlNet-XL} & 32.5 & 79.3 & \textbf{0.298} & \textbf{0.296} & \textbf{1.17} & \textbf{5.99} & \textbf{2.65} \\
        \bottomrule
    \end{tabular}
\end{table*}

\textbf{Motivation.}
Our main experiments are conducted on Stable Diffusion v1.5 (SD-v1.5) for
practicality and reproducibility. To address the concern that SD-v1.5 may not
reflect the capability of modern text-to-image systems, we additionally evaluate
Composer on Stable Diffusion XL (SDXL), a substantially stronger base model with
improved aesthetic quality and generation fidelity.

\noindent\textbf{Backbone and adapter.}
We build \emph{Union-ControlNet-XL} on top of SDXL-base~\cite{podell2023sdxl}.
Compared to the SD-v1.5 backbone, SDXL contains more attention blocks while
maintaining the UNet-style text conditioning pathway.
Our composition condition encoder (structure and color branches) follows the same
design as in SD-v1.5.

\noindent\textbf{Training setup.}
Since SDXL is trained at $1024\times1024$ resolution, we train Composer-XL on
JourneyDB~\cite{sun2023journeydb}.
To reduce compute while leveraging our curated $512$-resolution high-quality dataset,
we adopt a two-stage strategy:
(i) train at $512$ resolution for $200\text{k}$ steps with batch size $128$ to learn
composition-conditioned control at low resolution; then
(ii) continue training at $1024$ resolution for $50\text{k}$ steps with batch size $64$
to quickly adapt to high-resolution generation.
In practice, this saves roughly $\sim$50\% training compute compared to training
exclusively at $1024$ resolution.

\noindent\textbf{Baselines and protocol.}
We compare against representative SDXL-era control baselines, including
RPG-SDXL~\cite{yang2024mastering} and IP-Adapter-XL~\cite{ye2023ip}. We report text--image alignment (CLIP Score, HPS v2, ImageReward), image quality (FID),
and aesthetic score (Aes.) in Tab.~\ref{tab:comparison}. For fair comparison, IP-Adapter-XL uses the same
composition planner and reference pool as Composer-XL when a reference is required.

% ---------------------- SDXL qualitative figure ----------------------

\noindent\textbf{Parameter efficiency on SDXL.}
As shown in Tab.~\ref{tab:sdxl_ablation}, we further ablate the conditioning design on SDXL by comparing
(i) a \emph{Dual-ControlNet} (two independent ControlNet branches for structure and color)
and (ii) our \emph{Union-ControlNet-XL} (shared encoder with condition fusion and modulation).
Following Sec.~\ref{sec:ablation}, we report cycle-consistency losses for condition faithfulness
and also measure inference throughput (step/s) under the same hardware and sampling setup.

\section{Discussion}
\subsection{Generalization of Composition}
\label{sec:generalization_examples}

We provide representative \emph{generalization examples} to illustrate a central premise of our
design: the extracted composition conditions are intentionally \emph{low-information}
and \emph{semantic-agnostic}. Specifically, a single pair of conditions (coarse spatial
structure and global color distribution) can be reused to support diverse semantic
realizations, subjects, and abstraction levels. This suggests that visually complex
images can be grouped into a finite set of simple composition modes, where a
reference image serves as a representative instance of a mode rather than an exhaustive
semantic exemplar.

\begin{figure}[ht]
    \centering
    % Replace with your actual figure file (pdf/png)
    \includegraphics[width=\linewidth]{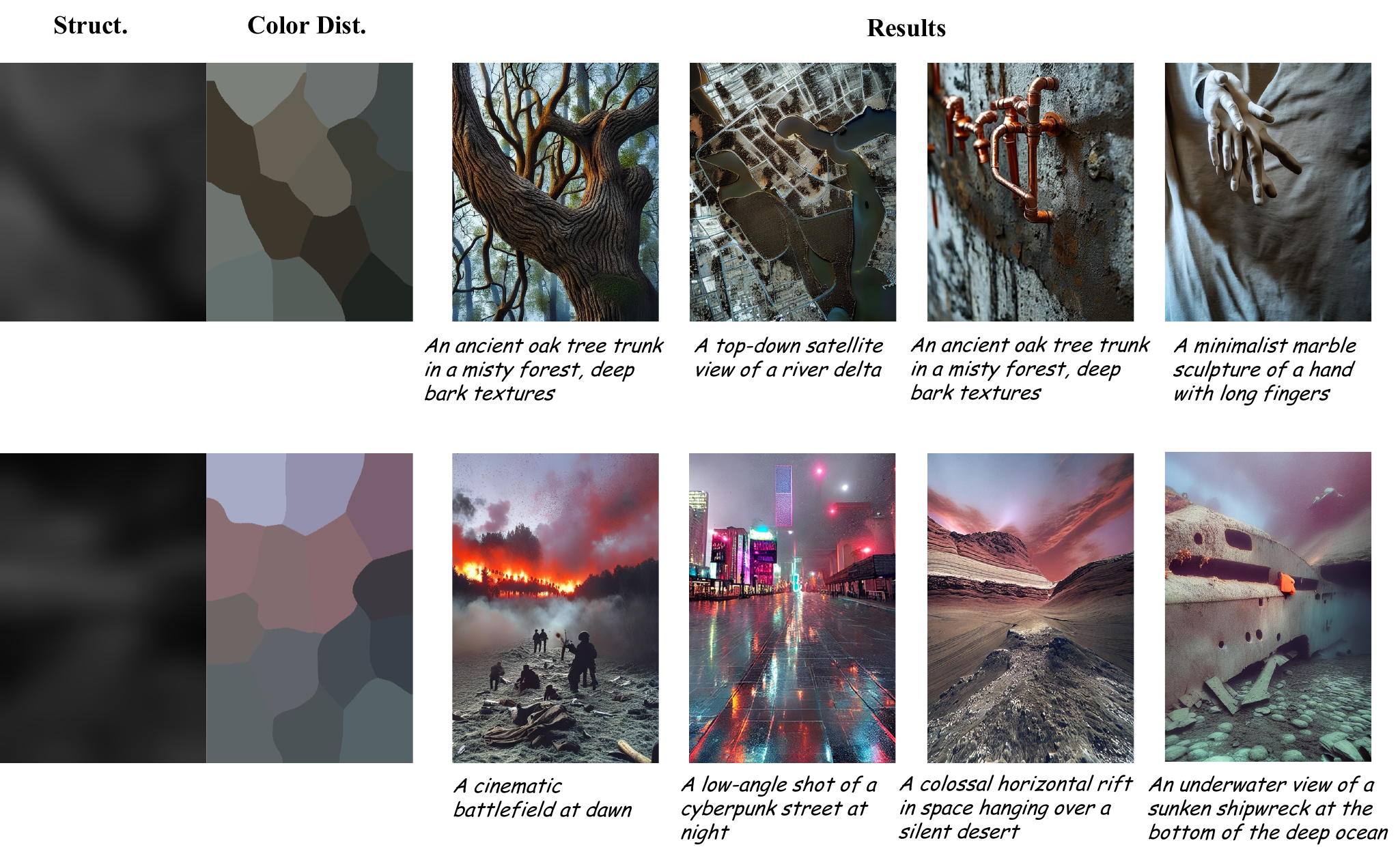}
    \caption{\textbf{One composition, many semantics.} Examples showing that the same
    extracted composition conditions (structure and color distribution) can support
    diverse semantic realizations across different scenes and abstraction levels.}
    \label{fig:generalization_examples}
\end{figure}

\subsection{Complementarity with Fine-grained Edge Control}
\label{subsec:fine_grained_edge_control}

\begin{figure}[h!t]
  \centering
  \includegraphics[width=.8\linewidth]{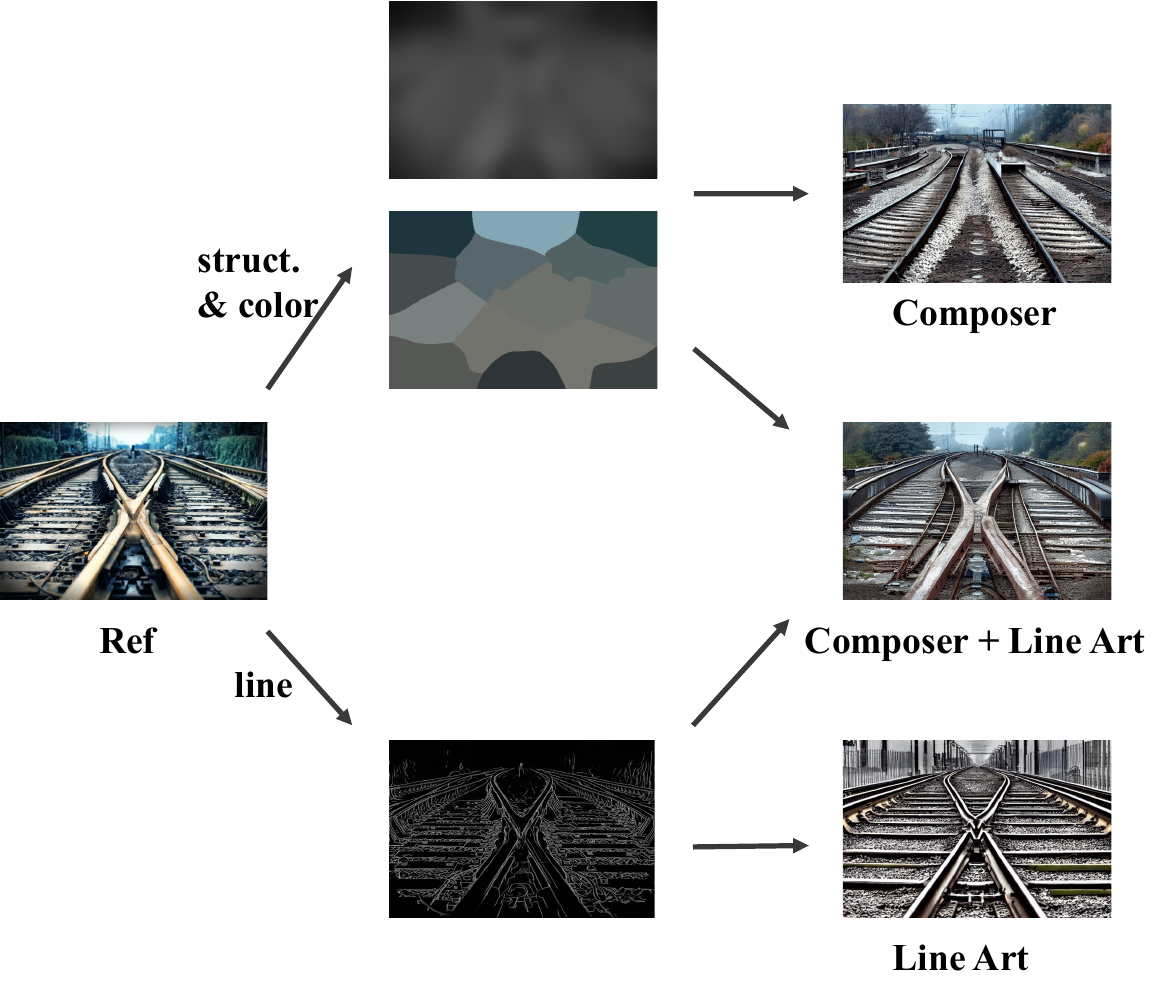}
  \caption{\textbf{Complementarity with fine-grained edge control.} Example results of combining our composition control (structure + color) with an off-the-shelf Lineart/Edge control. The composition control enforces the global layout and color distribution, while the edge/line control injects high-frequency geometric details (e.g., thin contours and framing lines), producing images that better preserve fine compositional aesthetics.}
  \label{fig:composition_lineart_combo}
\end{figure}

Our composition conditions intentionally focus on low-frequency cues---\emph{coarse} spatial structure and color distribution---to provide a semantic-agnostic global prior. As a result, high-frequency geometric details (e.g., thin lines, frames) may not be fully preserved. Importantly, this does \emph{not} preclude fine-grained control: since our Union-ControlNet modulates the backbone in a residual manner, it can be naturally \emph{composed} with off-the-shelf edge/line-based controls (e.g., Lineart/Edge ControlNet) at inference time without additional training. Fig.~\ref{fig:composition_lineart_combo} visualizes this synergy.

\section{Conclusion}
In this work, we introduce Composer, a novel framework for explicit modeling and control of image composition. It leverages spatial structure and color distribution representations, enabled by a tailored conditioning module trained on 2 million high-quality image-text pairs. Additionally, Composer performs theme‑driven composition planning: a Large Vision–Language Model (LVLM) retrieves reference images whose compositional cues augment the prompt. To go further, we fine‑tune the conditioning module with a text‑to‑composition objective, endowing it with implicit composition planning that infers compositional guidance directly from text. Experiments show that Composer  not only enhances the aesthetic quality of generated images but also supports flexible condition combinations and manipulation, offering users precision and flexibility in the creative process.

\section*{Statements and Declarations}

\noindent\textbf{Data Availability Statement.} 
The 2 M image–text pairs used to train \emph{Composer} are openly available at
\href{https://huggingface.co/datasets/jackyhate/text-to-image-2M}{\texttt{huggingface/text-to-image-2M}}
under an MIT licence.  
A citable snapshot of the dataset is archived on Hugging Face with the DOI
\href{https://doi.org/10.57967/hf/3066}{10.57967/hf/3066}.

% -------------------------------------------------
% (Add Funding / Competing Interests statements here if needed)
% -------------------------------------------------


\bibliography{sn-bibliography}% common bib file

\end{document}